\pdfoutput=1
\documentclass[11pt]{article}

\usepackage{ACL2023}

\usepackage{times}
\usepackage{latexsym}
\usepackage{multirow}
\usepackage{overpic}

\usepackage[T1]{fontenc}
\usepackage{booktabs}
\usepackage[utf8]{inputenc}

\usepackage{microtype}

\usepackage{inconsolata}
\usepackage{graphicx}

\usepackage{amssymb,paralist}

\makeatletter
\newcommand{\printfnsymbol}[1]{%
  \textsuperscript{\@fnsymbol{#1}}%
}

\title{MetaVL: Transferring In-Context Learning Ability \\ From Language Models to Vision-Language Models}

\author{Masoud Monajatipoor \\
  UCLA \\
  \texttt{\normalsize{monajati@ucla.edu}} \\\And
  Liunian Harold Li \thanks{equal contribution} \\
  UCLA \\
  \texttt{\normalsize{liunian.harold.li@cs.ucla.edu}} \\\And
  Mozhdeh Rouhsedaghat \printfnsymbol{1} \\
  USC \\
  \texttt{\normalsize{rouhseda@usc.edu}} \\
  \AND
  Lin F. Yang  \\
  UCLA \\
  \texttt{\normalsize{linyang@ee.ucla.edu}} \\\And
  Kai-Wei Chang  \\
  UCLA \\
  \texttt{\normalsize{kwchang@cs.ucla.edu}} \\
  }

\begin{document}
\maketitle
\begin{abstract}
%our proposed method 
%In computer vision, 

Large-scale language models have shown the ability to adapt to a new task via conditioning on a few demonstrations (i.e., in-context learning). % Meta-learning, a game-changing technique in low-resource settings, is shown to further improve their in-context learning ability. %make it a stronger few-shot learner. %specially for in-context learning. 
%However, in the vision-language domain, in-context learning is still considered a challenging problem and several attempts have been made to transfer language model in-context learning ability to multimodality. 
Large-scale language models have shown the ability to adapt to a new task via conditioning on a few demonstrations (i.e., in-context learning).
However, in the vision-language domain, most large-scale pre-trained vision-language (VL) models do not possess the ability to conduct in-context learning. How can we enable in-context learning for VL models? In this paper, we study an interesting hypothesis: can we transfer the in-context learning ability from the language domain to the VL domain? Specifically, we first meta-trains a language model to perform in-context learning on NLP tasks (as in MetaICL); then we transfer this model to perform VL tasks by attaching a visual encoder. Our experiments suggest that indeed in-context learning ability can be transferred cross modalities: our model considerably improves the in-context learning capability on VL tasks and can even compensate for the size of the model significantly. 
On VQA, OK-VQA, and GQA, our method could outperform the baseline model while having $\sim$20 times fewer parameters.% and a $\sim$ 6 times smaller VL training set. %on a an unseen VL task.
%to overcome this problem, Leveraging meta-traning is a potential 
%. furthermore, meta-learning is less explored due to the lack of multimodal meta-training datasets and tasks. 
%In this study, we propose that meta-learning knowledge for in-context learning can be transferred from single-modality to multimodality. Our proposed Vision-and-language (VL) model, MetaVL, considerably improves the in-context learning capability in the target model and can even compensate for the size of the model significantly. 
%By conducting extensive experiments on %various VQA tasks including 
%VQA, OK-VQA, and GQA, we find that MetaVL could outperform the baseline in in-context learning even with a $\sim$ 20 times smaller model size.% and a $\sim$ 6 times smaller VL training set. %on a an unseen VL task.
%including in-context learning in which the model is conditioned on a few training samples for making a prediction. Unlike language models, the performance of large-scale Vision-and-language (VL) models trained with large-scale data 
%are trustworthy for various VL tasks, their performance 
%drops significantly in this few-shot scenario. %In this work, we explore recent progress in this line of research and 
%Although Meta-training in vision-only models is well studied, in multimodality is less explored due to the lack of meta-train datasets and tasks. In this work, we 
%Although larger language models could benefit VL models for in-context learning, 

\end{abstract}

\section{Introduction}
Pre-trained language models have shown impressive performance on a range of tasks by learning from large-scale text corpus
%is a learning method which enables learning from massive datasets for high-quality open-ended text generation 
\cite{radford2018improving,radford2019language,yang2019xlnet}.
%which has shown impressive performance in many natural language tasks.% Large-scale language models could be trained on massive datasets in an autoregressive manner for high-quality open-ended text generation. 
Recent studies find that some of these language models can be used to perform \textit{in-context learning} out-of-the-box, i.e., adapting to a task by conditioning on a few demonstrations in context without any gradient update~\cite{brown2020language, min-etal-2022-metaicl}, which is highly desirable.
% This learning strategy 
%has not only achieved tremendous success in solving the tasks in an open-ended manner but also
% has shown great potential to learn a new task by conditioning on a few examples in context without any gradient update (in-context learning) \cite{brown2020language, min-etal-2022-metaicl}. 

In VL modeling, in-context learning is less explored and only a handful of models are proposed to perform in-context learning mainly by limiting the amount of deviation of a pretrained large-scale language model from the language space and translating visual inputs to language embedding space. They either require a large capacity \cite{tsimpoukelli2021multimodal, alayrac2022flamingo} or a giant corpus consisting of in-context learning examples \cite{alayrac2022flamingo, liu2023visual, koh2023grounding}. 
%\harold{Maybe we can cite some recent papers such as Llava, Grounding Language Models to Images for Multimodal Generation}
%In VL modeling, while a similar learning paradigm is becoming a dominant approach due to its generative nature to solve the tasks like VQA in an open-ended manner \cite{wang2021simvlm,sollami2021multimodal}%While these models have a strong generative ability to solve VL tasks and have shown 
%, its performance for few-shot learning and generalization is limited. A recent line of research is exploring methods for VL models to learn a new task in-context with only few-shot data mainly by limiting the amount of deviation of a pretrained large-scale language encoder-decoder from the language space and translating visual inputs to language embedding space \cite{tsimpoukelli2021multimodal, eichenberg2021magma,yang2022empirical, alayrac2022flamingo}. However, these solutions are either using significantly large models or their performance is not satisfactory. %far behind supervised fine-tuning.
%Meta-learning is a data-driven strategy for \textit{learning to learn} from a few data examples and quickly adapting to a new task in a few-shot manner . 

In this work, we explore whether we could enable in-context learning in VL tasks without resorting to extreme scale-up. We study an interesting hypothesis: can we transfer the in-context learning ability from the language domain to the VL domain? To elaborate, not every language model exhibits excellent \textit{in-context} learning ability; recent studies~\cite{min-etal-2022-metaicl} show that one could explicitly train language models to perform in-context learning, by training the model on multiple tasks with in-context few-shot examples, a process that resembles meta-learning. Thus,  an intriguing query arises: when a language model is first meta-trained to perform in-context learning, can it be transferred to perform in-context learning for VL tasks better?

A remarkable observation in our study is the utilization of a meta-trained language model as the transformer encoder-decoder and the mapping of visual features to the language embedding space. This innovative approach led to the development of our proposed VL model (we name it MetaVL). Impressively, our experimental results demonstrate that MetaVL surpasses the baseline model's performance, even when MetaVL is designed to be 20 times smaller in size. %This intriguing finding highlights the potential efficacy and efficiency of our proposed methodology.

This study makes three main contributions: 1) To the best of our knowledge, this is the first attempt to transfer the meta-learning knowledge for in-context learning from single-modality to multimodality. 2) We propose a VL model, MetaVL\footnote{The code will be released soon}, which outperforms the baseline in in-context learning while having a much smaller model size. 3) Through extensive experiments on VQA, GQA and OK-VQA, we demonstrate the in-context learning capability of MetaVL and analyze its components.

\section{Related work}
\paragraph{In-context learning in VL.}
Frozen~\cite{tsimpoukelli2021multimodal} is the first attempt for in-context learning in multimodality by leveraging a frozen GPT-like language model as the language backbone and mapping visual features to the language embedding space. Frozen sheds light on the feasibility of benefiting from the frozen LMs in VL modeling to learn a new task from a few examples in context. 
MAGMA \cite{eichenberg2021magma} %follow the similar idea with slightly different model architecture and model size approaches with differnt
is another %work in this line of research that proposed an 
encoder-decoder architecture for VL pre-training which showed that adding adaptor blocks between the frozen language model layers %and using a more powerful vision encoder (CLIP) 
could further improve the performance for VL tasks in a few-shot scenario. %Compared with Frozen, in addition to being trained on an x8 larger set of VL datasets, MAGMA also includes the training splits of the target datasets to its training set, while Frozen is adapted to an unseen new task in-context (in-context learning). \harold{I think this detail about training tasks can be moved to experiment? Usually related work only gives broad overviews.}

Other recent works \cite{yang2022empirical, alayrac2022flamingo, zeng2022socratic} follow the similar principle as the previous works to tackle in-context learning in VL modeling %a similar in-context learning pattern as previous works 
and achieve superior results by leveraging extremely large-scale models. %GPT-3 with 175 billion and a 70 billion parameter VL model, x25 and x10 larger than Frozen, respectively. 
%~\harold{We should focus on the idea difference. PICA is very different from Flamingo. I suggest try to move Frozen, MAGMA, and Flamingo into one paragraph. For PICA, we could say: On the other hand, PICA  and socratic models ~\cite{socratic models from google} takes another route. Instead of learning to map visual embeddings into language model embedding space, they use an image captioner to generate captions and input the captions in the language model as discrete tokens. Such an approach maximally preserves the in-context learning ability of language models by transfer the VL task into a language-based task.}
%and a massive multimodal dataset, 
%and they could achieve superior results for in-context learning a new task.  
%Inspired by MAGMA, we leverage CLIP as our visual encoder to better translate from vision to language embedding space.
%the idea of using CLIP as a vision encoder instead of NFResnet plus visual prefix is shown to be a better translator from vision to language embedding space. therefore, we followed MAGMA and leveraged CLIP as our visual encoder.

In this paper, we study a problem overlooked in prior work: we delve into the possibility of enabling in-context learning for VL tasks without relying on extensive scalability. Our focus lies in exploring the hypothesis: Is it feasible to transfer the in-context learning capability from the language domain to the VL domain?
%\harold{We also need a short sentence saying how our work relates / differs from theirs: In this paper, we study a problem overlooked in prior work: [paraphrase our core research question; emphasize the scale-up point].}

\paragraph{Meta-learning in language modeling}
Large-scale language models have shown the capability to be trained on a new task if properly prompted with in-context examples, i.e., in-context learning. In this learning strategy, the language model is asked to generate the desired output, e.g., an answer in the question-answering task, which is prompted by a few data examples along with their corresponding supervision sampled from the training split, and the language model learns the task in context without performing any gradient updates.
%in an in-context learning way for a new task meaning that prompting a new data in new task by a few training data along with their label could let the model 
%learn the task in examples without performing any gradient updates on parameters. 
Although such training is highly data-efficient, its performance is far behind supervised fine-tuning. Therefore, inspired by \cite{vilalta2002perspective, evgeniou2004regularized,finn2017model, ruder2017overview}, MetaICL \cite{min-etal-2022-metaicl} proposes training the model for in-context learning as a kind of meta-learning.
%\harold{I changed the phrasing a bit. MetaICL does not really use traditionally mete-learning techniques. It is more about selling their idea as a kind of meta-learning.}. 
MetaICL meta-trained a gpt language model on a diverse set of natural language tasks and datasets and showed that meta-training a language model in an in-context learning manner could significantly improve the in-context learning capability of the language model for a new task. 
%In this work, we explore the feasibility of improving the in-context learning capability in a multimodal setting leveraging a uni-modal meta-trained language model.

\section{Approach}
In this section, we first explain the existing meta-training procedure for language modeling and then introduce our proposed method for in-context learning in VL.

\begin{figure}
%\vspace{-5pt}
\includegraphics[scale=0.34]{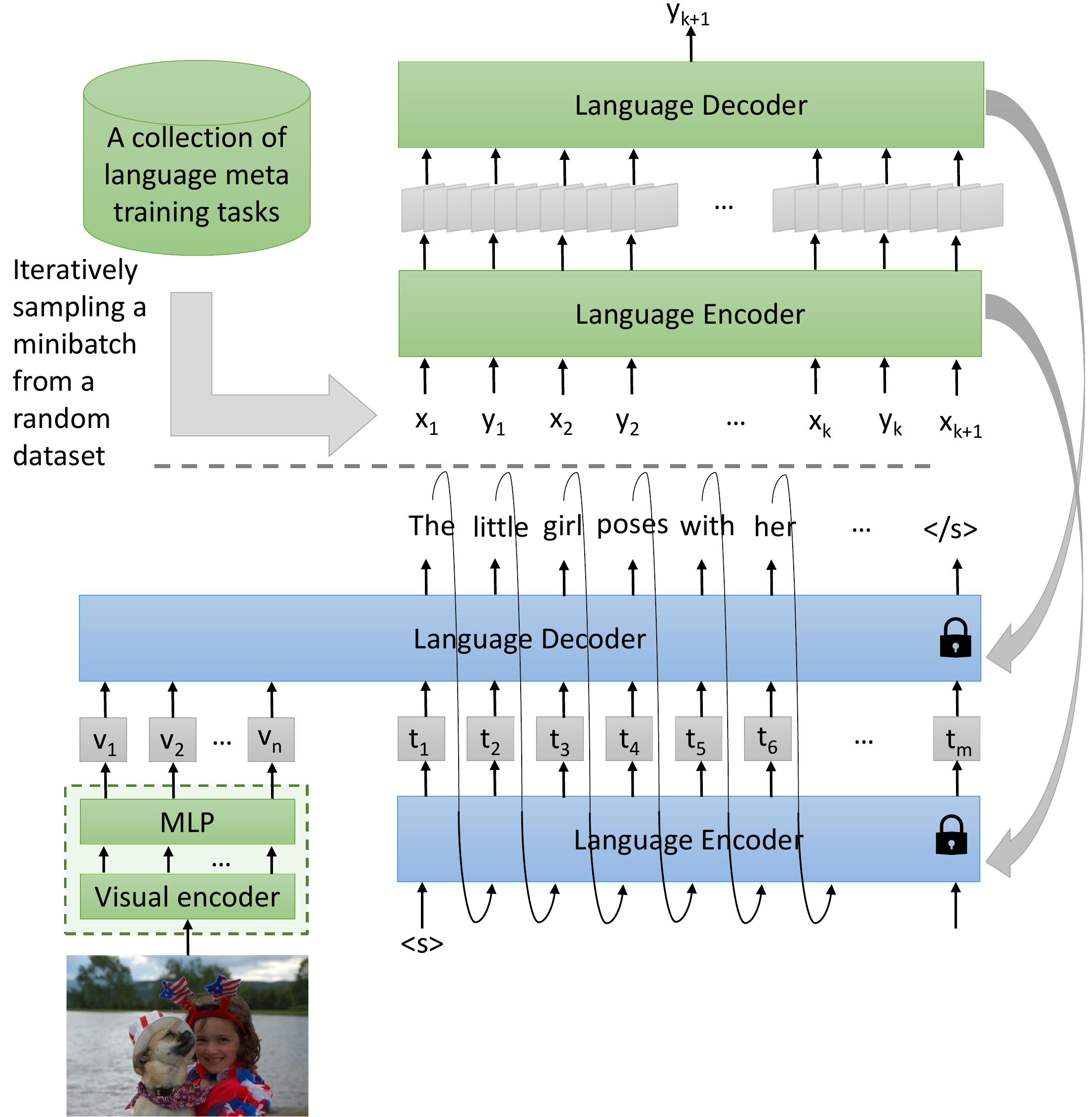}
\caption{The training steps of MetaVL including meta-training the language encoder-decoder (above) and mapping the visual features into the language embedding space while keeping the meta-trained language encoder-decoder frozen (below).} \label{fig1}
%\vspace{-5pt}
\end{figure}

\paragraph{Meta-training in language modeling.}

MetaICL has shown that a language model that is meta-trained on a diverse set of tasks in an in-context learning setup %meaning that a collection of k examples are given to the model in-context in each iteration 
is a strong few-shot learner.
%learning at inference (meta-learning) is a better few-shot learner for in-context learning on an unseen task. 
To meta-train an auto-regressive language model, in each iteration, a meta-learning task is randomly chosen from a collection of diverse meta-training language tasks, and $k+1$ data-label examples are randomly sampled from its training split. Then, the model is supervised by the concatenation of $(x_1,y_1, x_2,y_2, ... ,x_{k+1})$ %known as the support set 
which will be fed as a single input to the model for predicting the label $(y_{k+1})$ as the training objective%and predicting the label $(y_{k+1})$ will be the training objective.
%used and the model 
%Meta-training of LM is done by 
%randomly sample k+1 examples from the training split of a random task and concatenate k examples $(x_1,y_1) (x_2,y_2) ... (x_{k+1})$ which will be viewed as a single input and predicting the label $(y_{k+1})$ will be the training objective
%giving it to the model for language modeling objective training and let 
%ask the model to predict $(y_{k+1})$. 
, i.e., the meta-training step aims to maximize: 
%\vspace{-5pt}
\begin{equation}
P(y_{k+1}|x_1, y_1, \cdot\cdot\cdot , x_k, y_k, x_{k+1})
\end{equation}

During inference, the same in-context setup ($k$ examples from the training) %and $1$ from validation/test set) 
are sampled from a target dataset to be used as the $(x_1,y_1) (x_2,y_2) \cdot\cdot\cdot ,(x_k,y_k) (x)$ and given to the model to predict the label $y$. 

The meta-trained language model trained on a diverse set of natural language datasets has shown good performance for an unseen task when few data are given in context \cite{min-etal-2022-metaicl}.

%%if we formulate the visual inputs in the language embedding space.... 
%We call our model MetaVL that similar to Frozen is an encoder-decoder structure however the gpt model was meta-trained the same as MetaICL.

%%Frozen utilizes a GPT-like language model with 7B parameters while we trained metaVL with a much smaller language model (GPT2-medium) with 345M parameters that has been meta-trained. 

\paragraph{MetaVL - a VL method with meta-learning knowledge for in-context learning.} 
MetaVL has three main submodels including a meta-trained encoder-decoder and is being trained using Prefix Language Modeling (PrefixLM) \cite{wang2021simvlm}. 
 In the following, we discuss each submodel in detail.

\paragraph{Visual encoder and visual prefix.}
The visual encoder is %the first submodel of the multimodal encoder that can be 
defined as a function $V_{e}(x)$ that takes an image of x and outputs visual features. We extract the feature grid before the pooling layer $n \times D_v$ where $n$ is the number of feature maps and $D_v$ is the feature size of the visual encoder. Then, the output features can be viewed as a sequence of $n$ visual tokens representing the image.

%The first submodule in the multimodal encoder is the vision encoder. 
%For the vision encoder, we leverage CLIP which is a stronger vision encoder compared with the Frozen visual encoder -ResNet. We extract the feature grid before the pooling layer (144 in CLIP-RN50x16). So, the image is represented as a sequence of 144 image features that can be seen as 144 tokens describing the image. 

The visual encoder is followed by the visual prefix module that is defined as $V_{p}(x) \in D_v \times D_l $ which maps the visual features to language embedding space. This module is seeking to properly project the visual tokens into language tokens.

During the VL training, the parameters of both of these modules are trainable and are learned with different learning rates by back-propagation guided by the frozen language model.

\paragraph{Language encoder-decoder}
The meta-trained language encoder-decoder is used as the LM backbone and is frozen during the VL training process so the meta-trained language model preserves its few-shot capabilities. The language encoder encodes the text into text tokens represented by $t_1, t_2, ..., t_m$.
Then, given the multimodal tokens (image and text) as 
$U = {{v_1,v_2, ..., v_n, t_1, t_2, ..., t_m}} $ %The encoder is a multimodal encoder that takes and image-text pairs and the model decoder 
the decoder is trained to reconstruct the corresponding text with a standard language modeling objective to maximize the following likelihood:
\vspace{-5pt}
\begin{equation}
L(U) = \sum_{i=1}^{m} \log P(t_i|v_1,...,v_n,t_1,... t_{i - 1}; \theta)
\end{equation}
After the VL training, for learning a new VL task in-context, given a few examples from a new task with a new format, we concatenate k sampled data-label pairs from the training split along with one data from the val/test split to construct the prompt and feed it to the model for predicting the desired output. The entire process is visualized in Fig. \ref{fig1}.

\section{Experiments}

%& 38.2
%& 12.6

%\paragraph{Datasets}
\subsection{Datasets and Baseline}
We use the dataset proposed in \cite{min-etal-2022-metaicl} as the meta-training dataset for the language model and the COCO dataset \cite{lin2014microsoft} as the VL training dataset for MetaVL. The evaluation experiments are conducted on three datasets including VQA \cite{antol2015vqa}, OK-VQA \cite{marino2019ok}, and GQA \cite{hudson2018gqa}.
Frozen leveraged an internal GPT-like language model with 7 billion parameters as the backbone of their proposed model. %and could achieve encouraging results on VL tasks 
As their model is not publicly available, we trained Frozen with GPT2-Medium as the frozen language model and consider it as our main baseline (Frozen$_{\mathrm{A}}$) due to its model size. We also train a frozen with GPT-J 6B (The most similar GPT to Frozen) language model and obtained a close performance to the original Frozen model and use it as our second baseline denoted by Frozen$_{\mathrm{B}}$.
%reproduced Frozen with GPT-j-6B language model (Frozen\_B). 
%Their results are all reported in the Table \ref{main_result} along with results in \cite{tsimpoukelli2021multimodal} denoted by Frozen$_{\mathrm{ori}}$. 

\subsection{Training and evaluation setting}

\begin{figure}[ht]
\begin{center}
\centering
\includegraphics[scale=0.165]{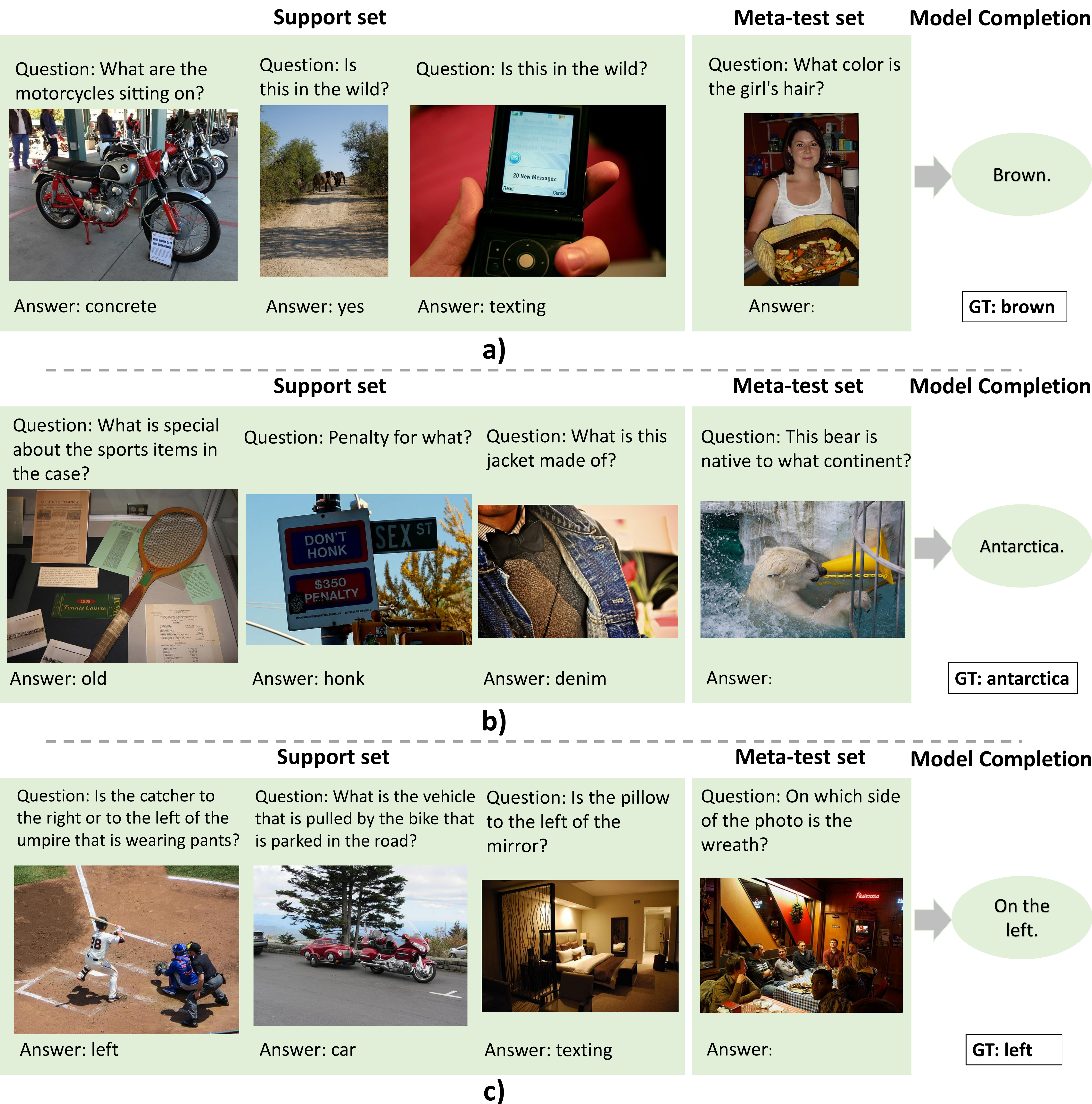}
\caption{Qualitative examples of in-context learning from three datasets: a) VQA, b) OK-VQA, and c) GQA. For each example, there is also a task induction sentence of ``please answer the question.''.}
\label{fig2}
\vspace{-5pt}
\end{center}
\end{figure}

%We meta-trained a GPT2-Medium on a collection of 142 meta-training language datasets. 
%In the following sections, we discusse how the meta-learning atribute is being transfered to multimodality and our model is a step toward transfering this capibility to multimodality learning (VL).
Initially, We meta-train a GPT2-Medium LM on a collection of 142 meta-training language datasets with a learning rate of 1e-5 and a batch size of 8 using the setting named as ``HR$\rightarrow$LR with instructions (all)'' where datasets with equal or greater than 10,000 training examples are used as meta-training tasks and the rest of the datasets are used as target tasks. The training is done on 8 NVIDIA RTX A6000 for 80,000 steps which took $\sim$ 6 hours. %Also, followed by
Then, we train MetaVL on the training split of COCO where we use a learning rate of 5e-5 and 2e-6 for the visual prefix and visual encoder, respectively, while the rest of the model parameters are frozen. We use a batch size of 32 and trained MetaVL using 4 NVIDIA RTX A6000 for 8 epochs which take $\sim$ 48 hours. Inference time depends on the numebr of shots varies from 2-5 hours for 0-3 shots on 5000 test examples. Our visual encoder is CLIP-RN50x16 \cite{radford2021learning} with a feature grid size of $144 \times 3072$ and our visual prefix is an MLP layer with a dimension of $3072 \times 768$.
For in-context evaluation on VQA datasets, we randomly pick a specific number -n- of sampled data-label pairs, known as shots, from the training set and feed them to the model in-context followed by a single data from the val/test set. Fig. \ref{fig2} provides some illustrative examples for the evaluation process.

%For evaluation, we get the output of the model and check if it exactly match the answer. if not, then we pick the most similar answer in the set of candidate answers for each dataset to the generated output as model output and check its match with the corresponding answer. VQA, OK-VQA, and GQA have ~3000, ~4200, and ~3000 unique answers in the training set. We also did another round of human evaluation summarized in Table 2. 

%For evaluation, due to limited computation, on 5000-size sample data from the val/test dataset, we compare the output generated by the model against the expected answer following prior works. If they do not match exactly, we select the most similar answer from a set of candidate answers as the model's output (We use the Cosine similarity between the embedding of the output and each candidate answer using sentence BERT).
To conduct the evaluation, we utilize a subset of 5,000 instances from the val/test dataset due to computational constraints. The generated output from the model is then compared against the expected answer, as established in previous studies. In cases where an exact match is not achieved, we employ a technique to identify the most closely related answer from a set of candidate answers (The set can be defined as a unique list of all answers in the training dataset). This involves computing the cosine similarity between the output's embedding and each candidate answer's embedding achieved by Sentence BERT \cite{reimers-2019-sentence-bert}.

\begin{figure}[ht]
\begin{center}
\centering
\includegraphics[scale=0.25]{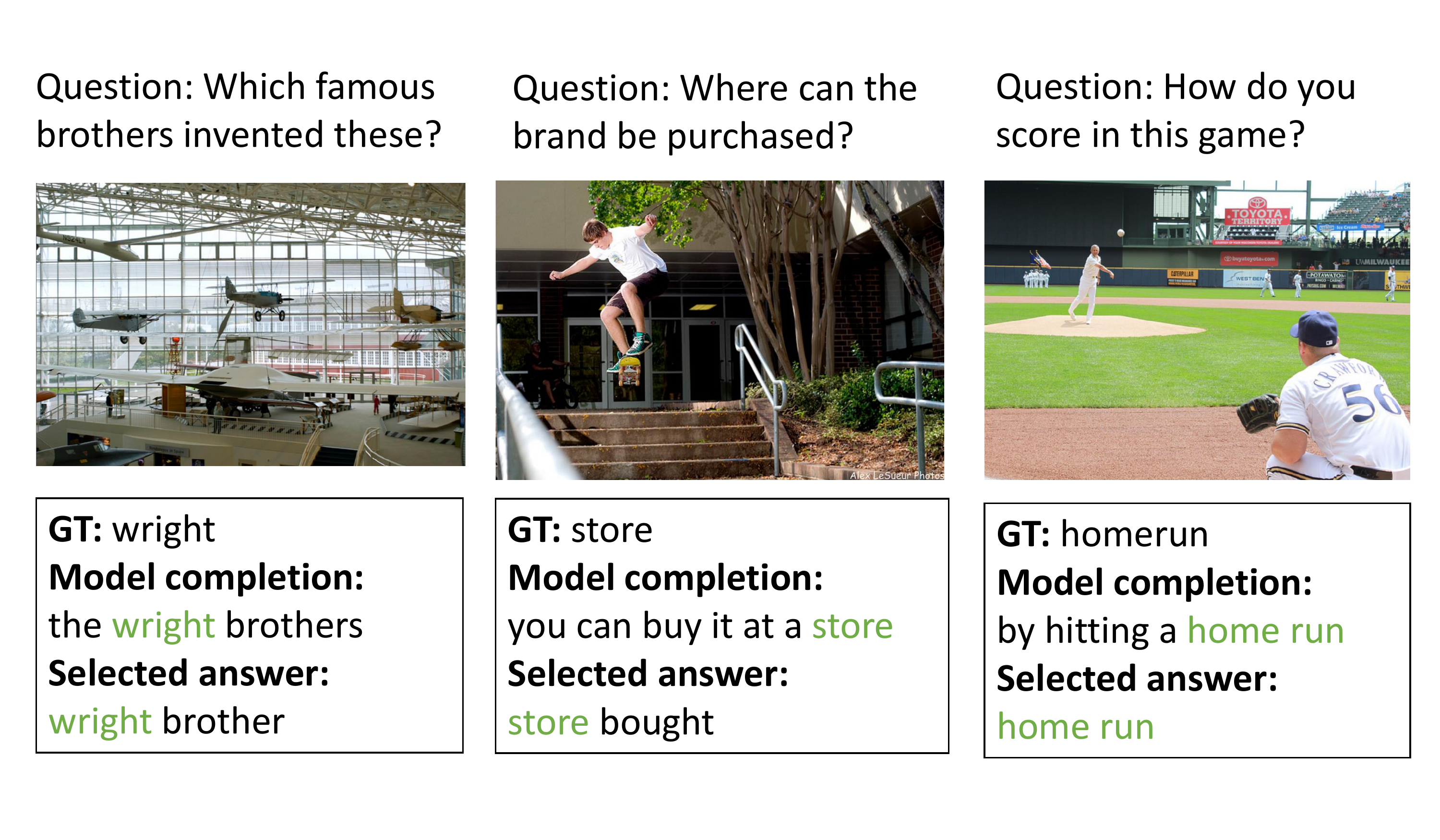}
\caption{Three examples of VQA cases which The model's output, although correct, slightly differs from the ground-truth and selected answer from the candidate set.}
\label{fig3}
\vspace{-5pt}
\end{center}
\end{figure}

We then compare the selected output with the corresponding answer to determine the match. The training datasets for VQA, OK-VQA, and GQA contain approximately 3,000, 4,200, and 3,000 distinct answers, respectively. Furthermore, we performed an additional round of human evaluation on model's output without matching, and the findings are summarized in the appendix (Table 2). The human evaluation on a separate test set of 2000 examples aimed to delve deeper into instances where the model's output, while accurate, didn't precisely match the provided answer. Three such examples are presented in Fig \ref{fig3}, where the initial evaluation did not consider the prediction as correct, but it was deemed correct in the subsequent evaluation setting.

\subsection{Results and analysis}

\paragraph{Quantitative analysis}
To evaluate MetaVL, we consider three common visual question-answering datasets including VQA, OK-VQA, and GQA.
%and evaluated all settings with different 
We compare MetaVL results with the mentioned two baselines %and also the original Frozen results, shown by Frozen$_{\mathrm{ori}}$, 
in Table \ref{main_result} for 3-shot in-context learning based on both automatic and human evaluation.
%three variants of Frozen: Frozen\_A: the results of Frozen in \cite{tsimpoukelli2021multimodal}, Frozen\_B: the results of reproducing \cite{tsimpoukelli2021multimodal} given GPT-J 6B language model, Frozen\_C: the results of reproducing \cite{tsimpoukelli2021multimodal} given GPT2-Medium language model.
According to the results, the performance of Frozen improves as its model size increases while MetaVL achieved competitive results in all three tasks. To further analyze how many image-text pairs are required to enable In-context learning for the VL task, we have trained MetaVl with 50 percent of training data and the results show that the performance slightly dropped but the model preserve its capability to learn from in-context data (Table \ref{table6}). %Please note that although the original 
%Compared to the frozen results on VQA and OK-VQA, 
\begin{table}
\small
\begin{center}
  \centering
  \resizebox{\linewidth}{!}{
  \begin{tabular}{ccccc}
  %\hline
    \toprule
    %\multicolumn{3}{c}{Part}                   \\
    %\cmidrule(r){2-5}
     & & Frozen$_{\mathrm{A}}$ & Frozen$_{\mathrm{B}}$ &  MetaVL\\
    %\hline\hline
    \midrule
    & LM size & ~375M & ~7B & ~375M\\ 
    \midrule
    \multirow{3}{*}{Automatic evaluation} & VQA  & 18.63 & \textbf{34.07} & 33.12 \\
    & OK-VQA  & 3.17 & \textbf{11.97} &  9.60 \\
    & GQA  & 13.86  & 25.76 & \textbf{31.96}  \\
    \midrule
    \multirow{3}{*}{Human evaluation} & VQA  & 16.68 & - & \textbf{ 35.09} \\
    & OK-VQA  & 6.41 & - &  \textbf{19.22} \\
    & GQA  & 19.96  & - & \textbf{38.29}  \\
    %\hline
    %\hline

    \bottomrule
  \end{tabular}
  }
  %\vspace{5pt}
  \caption{The performance of MetaVL compared with two baselines on 3-shot in-context learning. We report the performance of our re-implemented Frozen models.}
  \label{main_result}
  \end{center}
\vspace{-1pt}
\end{table}

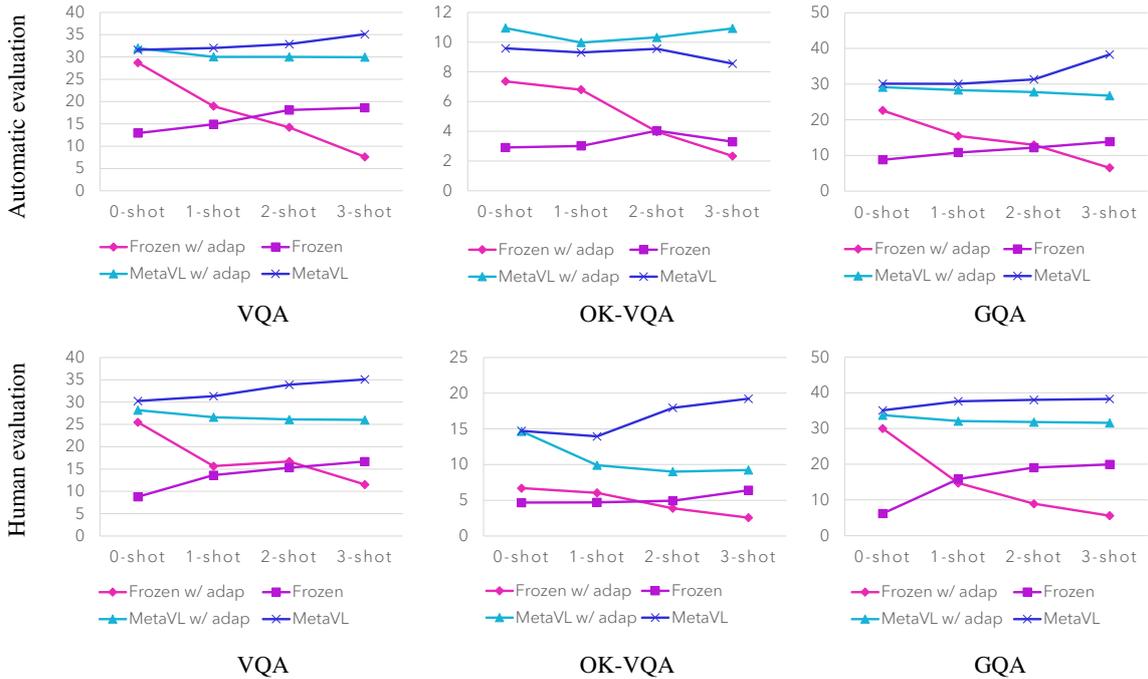
\begin{figure*}[ht]
%\begin{center}
\centering
\begin{overpic}
[width=0.94\textwidth]{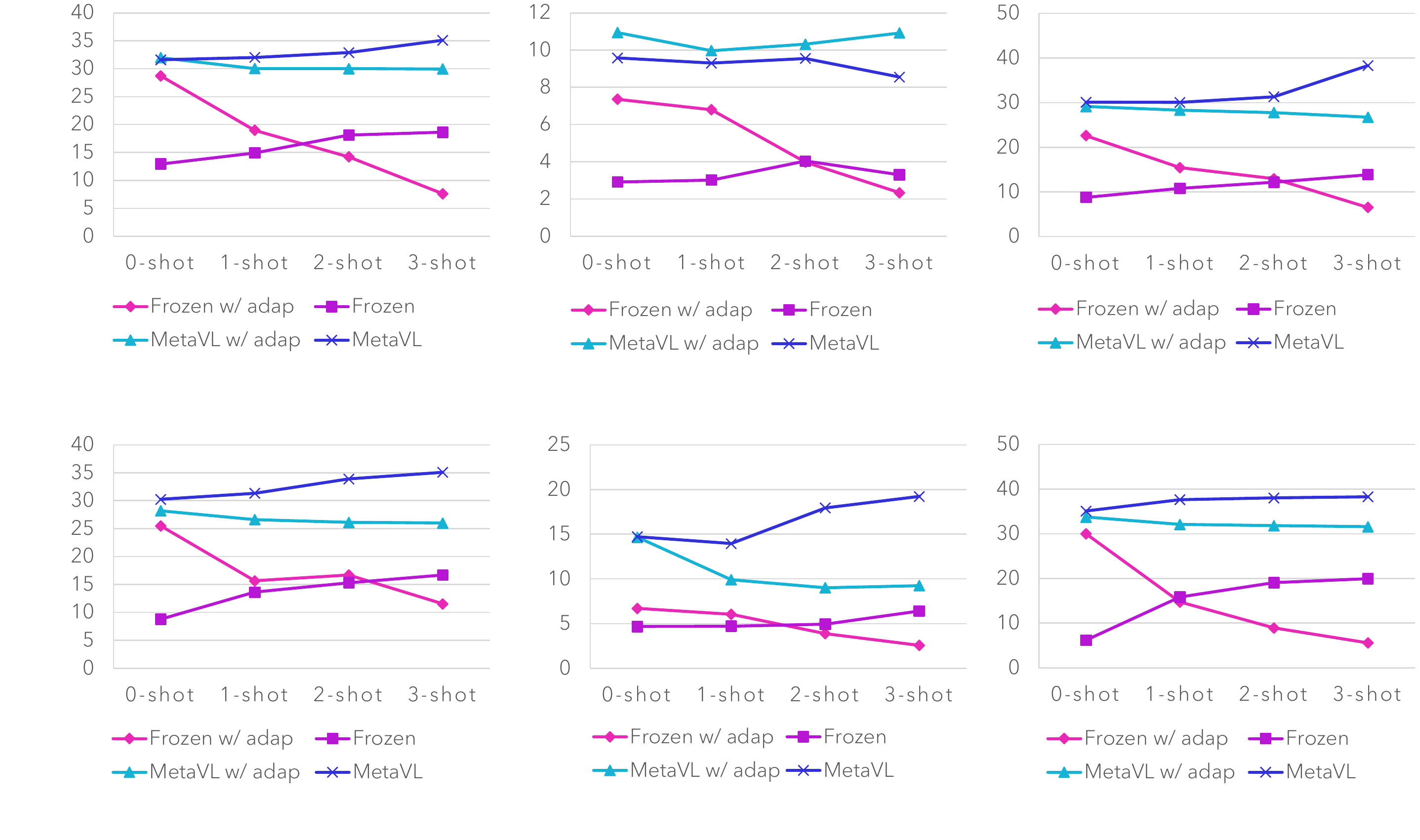}
    \put(0,40){\rotatebox{90}{\small{{Automatic evaluation}}}}
    \put(0,12){\rotatebox{90}{\small{{Human evaluation}}}}
    \put(20,31){\small{{VQA}}}
    \put(50,31){\small{{OK-VQA}}}
    \put(84.5,31){\small{{GQA}}}
    \put(20,0){\small{{VQA}}}
    \put(50,0){\small{{OK-VQA}}}
    \put(84.5,0){\small{{GQA}}}
\end{overpic}

\caption{Automatic and human evaluation Accuracy of MetaVL and Frozen, w/ and w/o adaptors with 0-3 shots of in-context data.}
\label{fig5}
%\vspace{-17pt}

%\end{center}
\end{figure*}

\paragraph{The effect of the number of in-context shots}
% demonstrates the effect of the number of shots given in context on the performance of different models. %how the performance of the model improves by increasing the number of shots given in context. 
According to Figure \ref{fig5}, in almost all settings, the performance of MetaVL is improving by increasing the number of shots which shows the model is gaining knowledge from the data in context. This result further gives us an illustration of the model's capability to learn from the in-context examples supporting that MetaVL is benefiting from the meta-learning knowledge for in-context learning. The numbers on the graph are summarized in Table \ref{table3} in the appendix.

\paragraph{The effect of having adaptor layers in LM}
MAGMA claims that adding trainable adaptor layers and letting the LM slightly be trained during the VL training process is beneficial for in-context learning. 
Compared with Frozen, in addition to being trained on an x8 larger set of VL datasets, MAGMA also includes the training splits of the target datasets to its training set, while Frozen is adapted to an unseen new task in-context (in-context learning).
We evaluated this method by adding adaptor layers to both Frozen and MetaVL and denoted the corresponding models by Frozen w/adap and MetaVL w/adap, respectively, in Fig. \ref{fig5}. 
%we evaluated the effect of adding adaptor layers to both Frozen and MetaVL and denoted 
%Following MAGMA \cite{eichenberg2021magma}, we added adaptor layers to MetaVL and Frozen to have new settings as MetaVL w/adap and Frozen w/adap respectively. 
Our results demonstrate that having a fully frozen language model in MetaVL could better preserve the in-context learning ability of the language model.
It is also noticeable that adding adaptor layers improves %comparing Frozen and Frozen w/adap shows Frozen w/adap has
the zero-shot performance of Frozen. We hypothesize that this improvement is due to getting a better vision and language alignment by letting both vision and language submodels be involved in the alignment process.
%we argue that it moves the model toward a fully trainable language model that has better zero-shot \cite{wang2021simvlm} capability rather than few-shot capability in context. 
%MetaVL has shown better zero-shot and few-shpt performance compared with its setting with adaptors.
%\vspace{-5}

\paragraph{Qualitative analysis}
We provide some qualitative examples to better illustrate the performance of MetaVL for in-context learning in different VQA tasks. In Fig. \ref{fig2}, a few examples are provided which show the output of MetaVL for 3-shot in-context learning. More examples are presented in Appendix.

%To better elaborate in-context learning scenarios, During the inference for each question, we consider the question type and select in-context data among the similar category. For example, if the target question starts with "what color", we choose all in-context data starting with "what color". This could help the model better handle each question type. We also tried another experiment setting only randomly sampling in-context data and we report their results for both metaVL and baseline. In VQA, meta-trained language model with choosing in-context data from the same category beats all other settings. In OK-VQA and GQA, choosing from the same category doesn't make much different compared with randomly sampling. Our results are summarised in the table.

%For VQA task, our results for the 

%Our experiments and analysis on VQA demonstrate that meta-trained language model enriches the VL model for VL tasks in few-shot manner.
%The model is trained to generate the caption given an image.

%and evalauted the zero/few shot performance of the model on 

% \subsubsection{Impacts on }

\section{Conclusion}
We investigate the feasibility of transferring meta-learning knowledge for in-context learning from resource-rich single modality to 
%of transferring meta-learning knowledge for in-context learning from resource-rich single modality to 
multimodality. We have shown that by leveraging a meta-trained language model in a VL model, we can transfer the ability of ``learning to learn'' in context to VL and it results in a strong VL few-shot leaner. 
%we designed an strategy to leverage in-context learning ability of a meta-train language model and transfer it to multimodality 
%the transferability of the in-context learning ability of a meta-trained model from language to VL. 
%introduce a method for transferring the in-context learning ability of a meta-trained model from language to VL. 
With extensive experiments on three common VL datasets, we have shown that the in-context learning performance of MetaVL is superior compared with the baseline even when the size of our model is 20 times smaller.

%\section{Acknowledgment}This work is supported by funding 442521-YL-79453.
\section{acknowledgment}
This work was supported by DARPA under agreement HR00112190130 and DARPA MSC program under agreement N660011924032. We would like to thank the reviewers for their feedback to improve this research work.

\section*{Limitations}
While we have shown the potential of transferring in-context learning ability from a language model to VL tasks, the experiments in this paper are limited in two aspects. (1) We considered only the VQA task, which is limited in scope. It is unclear whether our method generalizes to other VL tasks. In fact, as most tasks in the VL domain take the form of visual question answering, it is less well-defined what would ``cross-task generalization'' entail in VL, compared to in NLP where (2) Due to computational limitations, we experiment with only a moderate-sized LM. It is unclear the performance of our method after scaling up. %In fact, as most tasks in the vision-language domain take the form of visual question answering, it is unclear what would ``cross-task generalization'' entail in VL, compared to 

\bibliography{anthology,custom}
\bibliographystyle{acl_natbib}

%\newpage
\appendix

\section{Appendix}
\label{sec:appendix}

\begin{figure*}
%\vspace{-100pt}
\includegraphics[width=\textwidth]{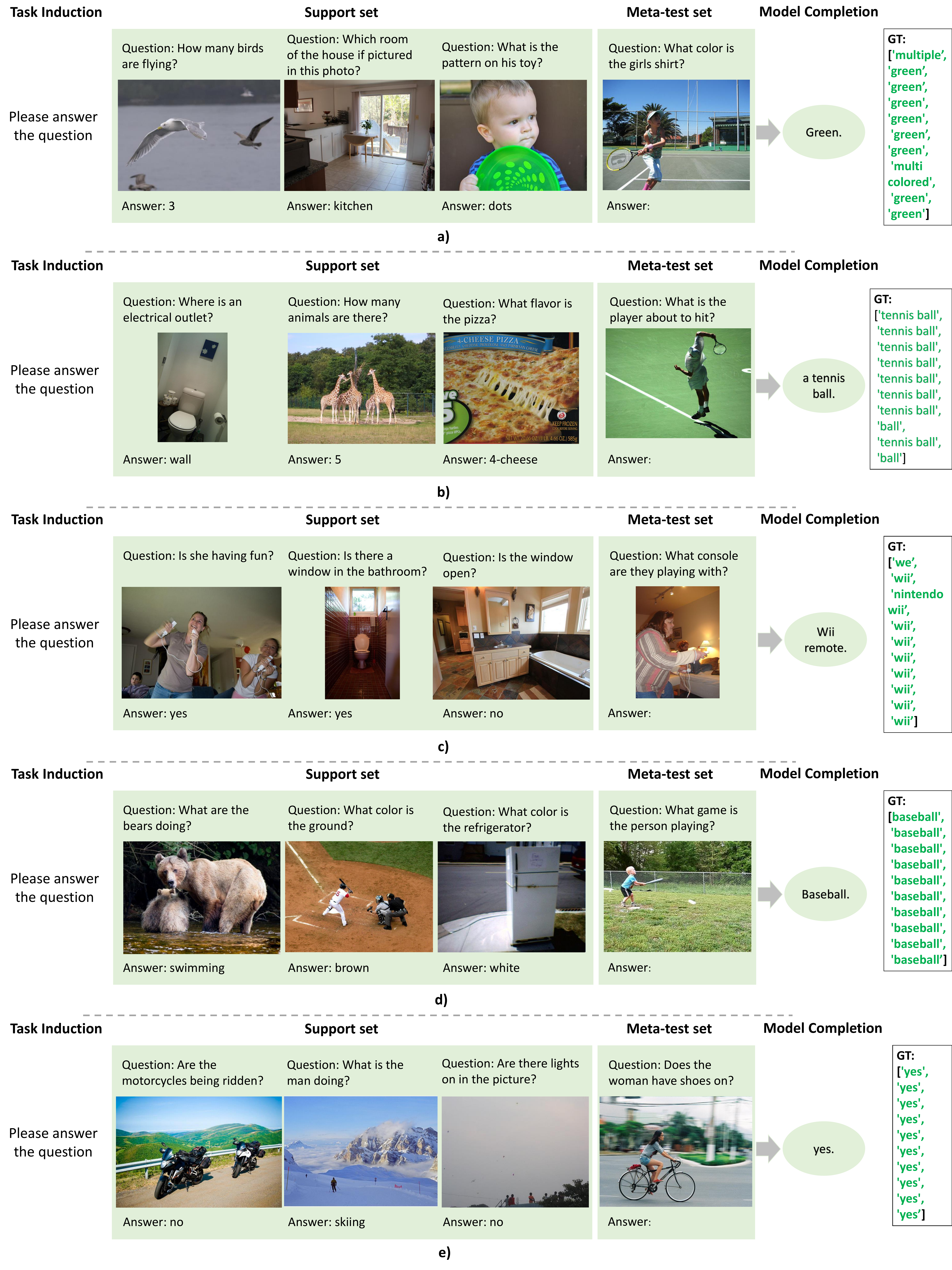}
\caption{MetaVL success examples from VQA.} \label{fig10}
\vspace{-15pt}
\end{figure*}

\begin{figure*}
\includegraphics[width=\textwidth]{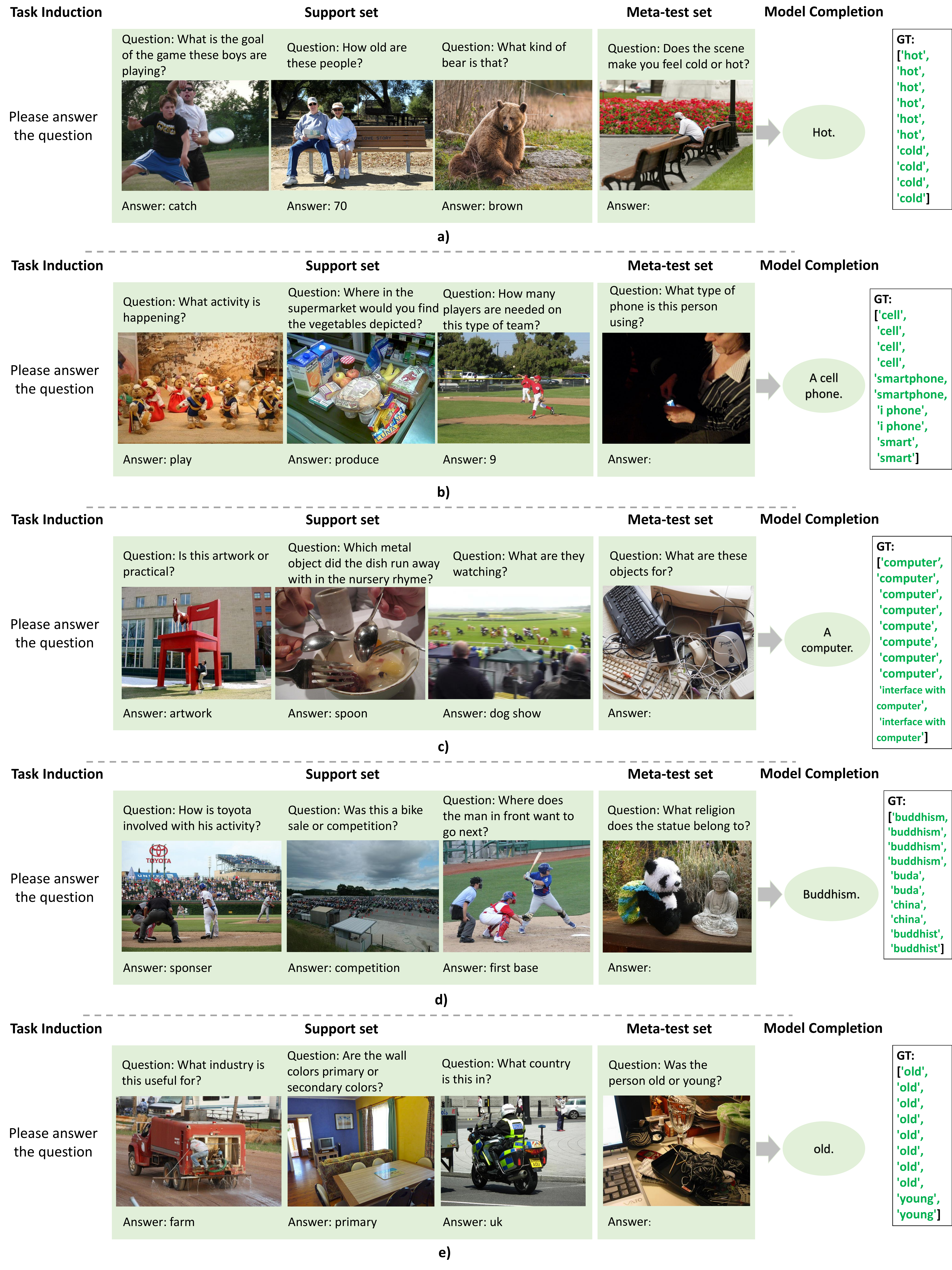}
\caption{MetaVL success examples from OK-VQA.} \label{fig11}
\vspace{-15pt}
\end{figure*}

\begin{figure*}
\includegraphics[width=\textwidth]{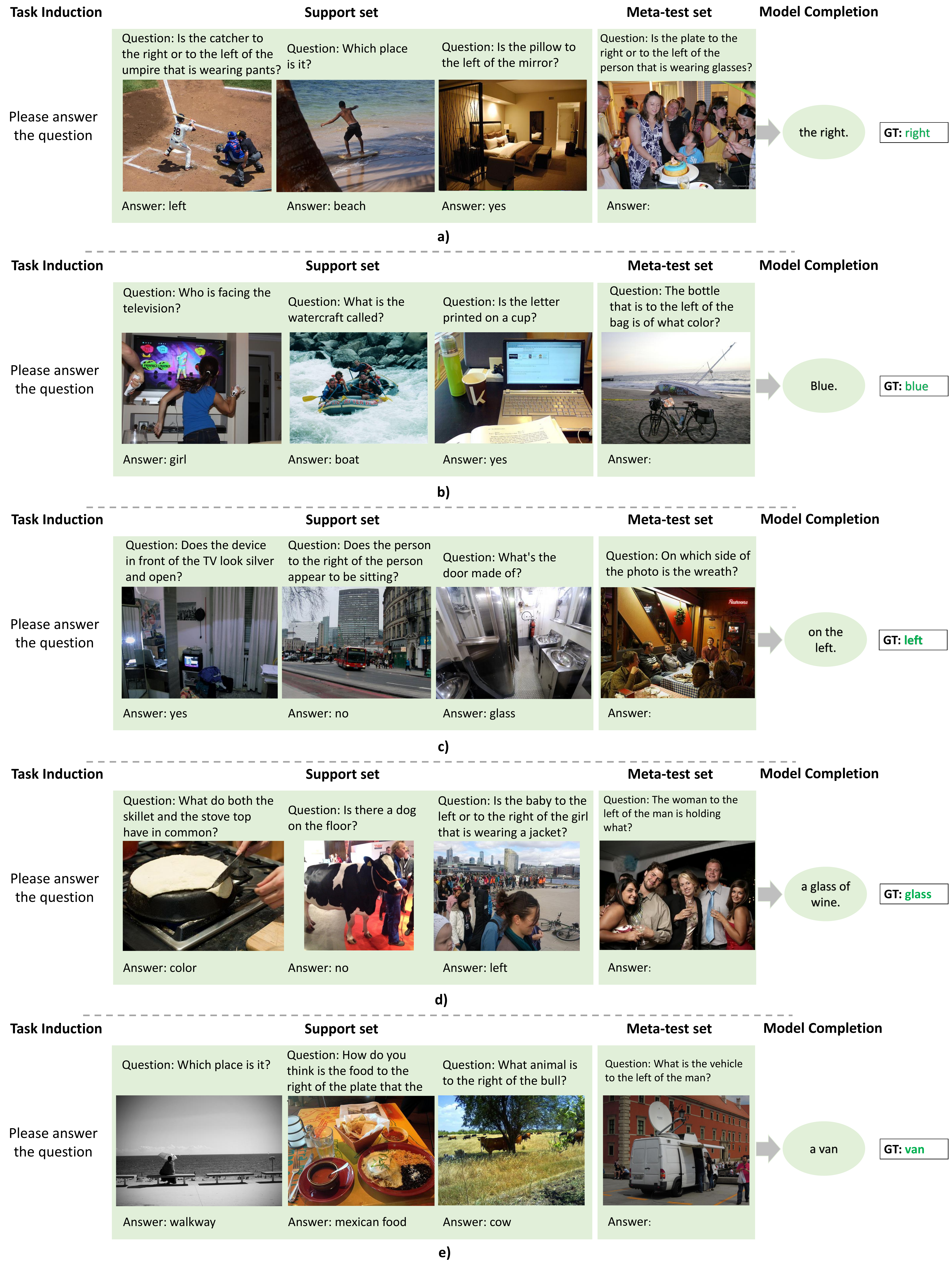}
\caption{MetaVL success examples from GQA.} \label{fig12}
\vspace{-15pt}
\end{figure*}

\begin{figure*}
\includegraphics[width=\textwidth]{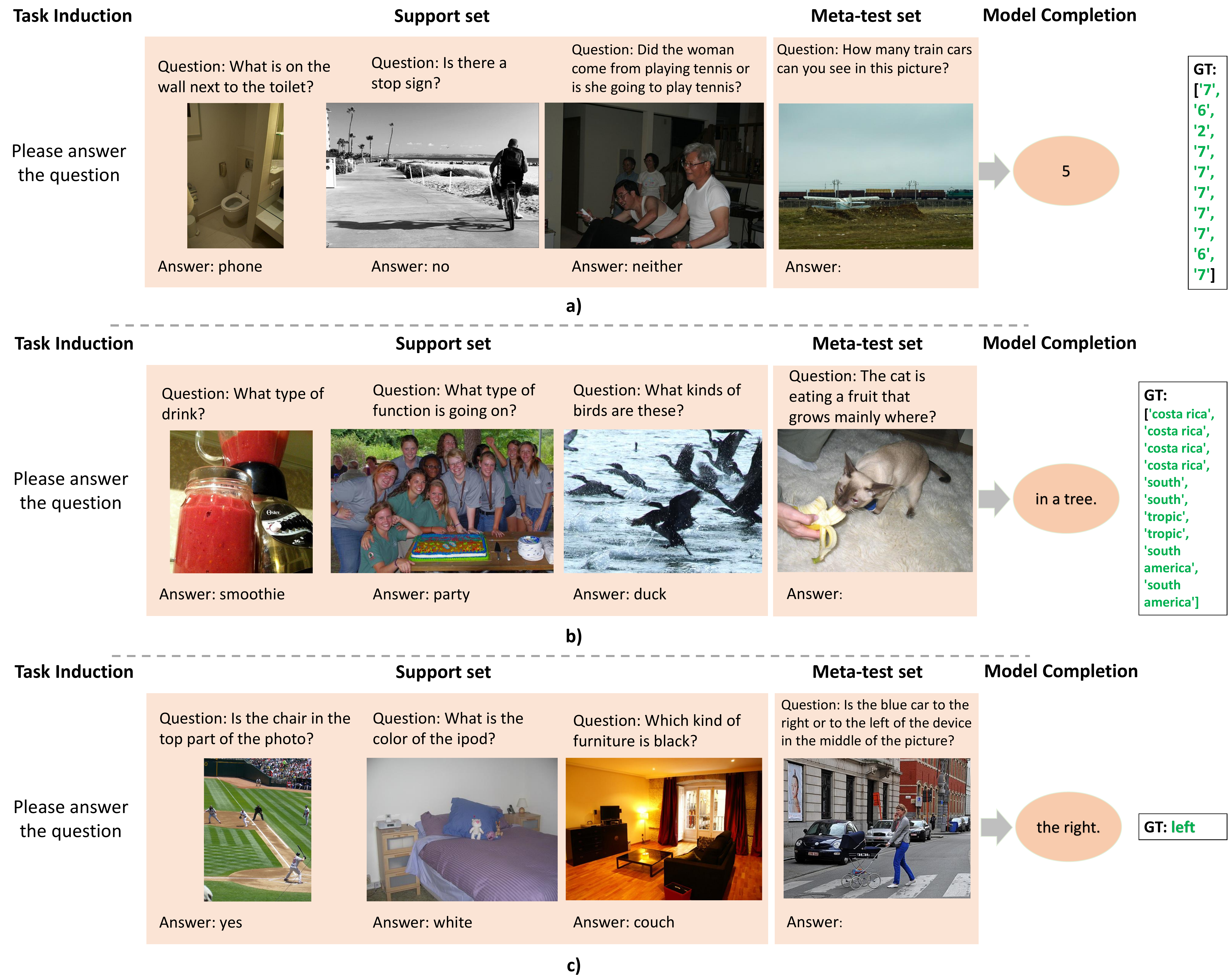}
\caption{MetaVL failure examples from a) VQA, b) OK-VQA, and c) GQA.} \label{fig13}
\vspace{-15pt}
\end{figure*}

\begin{table*}[ht]
\scriptsize
\begin{center}
  \centering
  \resizebox{\linewidth}{!}{
  \begin{tabular}{c|cccc|cccc|cccc|cccc}
  %\hline
    \toprule
    %\multicolumn{3}{c}{Part}                   \\
    %\cmidrule(r){2-5}
    model & \multicolumn{4}{c|}{Frozen$_{\mathrm{A}}$ w/ adap} & \multicolumn{4}{c|}{Frozen$_{\mathrm{A}}$} & \multicolumn{4}{c|}{MetaVL w/ adap} & \multicolumn{4}{c}{MetaVL} \\
    n-shot  & 0 & 1 & 2 & 3 & 0 & 1 & 2 & 3 & 0 & 1 & 2 & 3 & 0 & 1 & 2 & 3 \\
    %\hline\hline
    \midrule
    \multicolumn{17}{c}{Automatic evaluation} \\
    \midrule
    VQA & 28.72 & 18.98 & 14.23  & 7.60  & 12.94 & 14.92 & 18.11 & 18.63  &  31.98 & 30.03 & 30.01 & 29.96  & 31.6 & 32.01 & 32.89 & \textbf{33.12} \\
    OK-VQA & 7.36 & 6.30 & 3.98 & 2.34  & 2.91 & 3.02 & 4.04 & 3.30  & \textbf{10.94} & 9.97 & 10.32 & 10.92 & 9.58 & 9.30 & 9.55 & 9.60 \\
    GQA & 22.62 & 15.44 & 12.96 & 6.54 & 8.80 & 10.81 & 12.17 & 13.86  & 29.12 & 28.31 & 27.78 & 26.74  & 30.10 & 30.05 & 31.32 & \textbf{31.96}  \\

    \midrule
    \multicolumn{17}{c}{Human evaluation} \\
    \midrule
    VQA & 25.49 & 15.66 & 16.70  & 11.53  & 8.79 & 13.62 & 15.31 & 16.68  &  28.20 & 26.61 & 26.12 & 26.01  & 30.24 & 31.33 & 33.89 & \textbf{35.09} \\
    OK-VQA & 6.70 & 6.04 & 3.88 & 2.56  & 4.67 & 4.71 & 4.94 & 6.41  & 14.67 & 9.97 & 9.01 & 9.24 & 14.72 & 13.95 & 17.95 & \textbf{19.22} \\
    GQA & 30.01 & 14.72 & 8.92 & 5.59 & 6.18 & 15.85 & 19.07 & 19.96  & 33.74 & 32.09 & 31.81 & 31.58  & 35.08 & 37.65 & 38.03 & \textbf{38.29}  \\
    %\hline
    %\hline
    
    \bottomrule
  \end{tabular}
  }
  %\vspace{5pt}
  \caption{Accuracy of MetaVL and Frozen, w/ and w/o adaptors with 0-3 shots of in-context data.}
  \label{table3}
  \end{center}
\vspace{-1pt}
\end{table*}

\begin{table*}
\small
\begin{center}
  \centering
  \resizebox{\columnwidth}{!}{
  \begin{tabular}{cccc}
  %\hline
    \toprule
    %\multicolumn{3}{c}{Part}                   \\
    %\cmidrule(r){2-5}
     & & MetaVL & MetaVL$_{\mathrm{50\%}}$\\
    %\hline\hline
    \midrule
    \multirow{3}{*}{Automatic evaluation} & VQA & \textbf{33.12} & 30.32 \\
    & OK-VQA & \textbf{9.60} &  7.56 \\
    & GQA    & \textbf{31.96} & 27.77  \\
    \midrule
    \multirow{3}{*}{Human evaluation} & VQA   & \textbf{ 35.09} & 34.02 \\
    & OK-VQA  & \textbf{19.22} &  18.19 \\
    & GQA   & \textbf{38.29} &   35.66\\
    %\hline
    %\hline

    \bottomrule
  \end{tabular}
  }
  %\vspace{5pt}
  \caption{The performance of MetaVL was evaluated using the complete CoCo training dataset as well as a subset containing 50 percent of the CoCo training data. The experimental results indicate that even with the reduced training data, MetaVL maintains its capacity for in-context learning, albeit with a slight decrease in performance.}
  \label{table6}
  \end{center}
\vspace{-1pt}
\end{table*}

\end{document}